\pdfoutput=1

\documentclass[11pt]{article}

\usepackage[preprint]{acl}

\usepackage{longtable}
\usepackage{geometry}
\usepackage{graphicx}
\usepackage{adjustbox}
\usepackage{amssymb}
\usepackage{subfigure}
\usepackage{subcaption}
\usepackage{array,booktabs,ragged2e}
\usepackage{multirow}
\usepackage{multicol}
\usepackage{times}
\usepackage{latexsym}
\usepackage{arydshln}
\usepackage{hyperref}
\usepackage{pifont}
\usepackage{float}
\usepackage{bm}
\usepackage{microtype}
\usepackage{inconsolata}
\usepackage{xcolor}
\usepackage{xspace}

\usepackage{placeins}
\usepackage{afterpage}
\usepackage{amsfonts}
\usepackage{amsmath}

\usepackage[T1]{fontenc}
\usepackage[utf8]{inputenc}

\definecolor{up}{HTML}{D62F2F}
\definecolor{down}{HTML}{485694}

\newcommand{\diff}[1]{\small{\textcolor{up}{{(#1)}}}}
\newcommand{\improve}[1]{\small{\textcolor{down}{{(#1)}}}}

\title{Far Out: Evaluating Language Models on Slang in Australian and Indian English}

\author{
\textbf{Deniz Kaya Dilsiz}\quad
\textbf{Dipankar Srirag}\quad
\textbf{Aditya Joshi}\\
University of New South Wales, Sydney, Australia\\
\texttt{\{d.dilsiz, d.srirag, aditya.joshi\}@unsw.edu.au} 
}
\begin{document}
\maketitle
\begin{abstract}
Language models exhibit systematic performance gaps when processing text in non-standard language varieties, yet their ability to comprehend variety-specific slang remains underexplored for several languages. We present a comprehensive evaluation of slang awareness in Indian English (en-IN) and Australian English (en-AU) across seven state-of-the-art language models. We construct two complementary datasets: \textsc{web}, containing 377 web-sourced usage examples from Urban Dictionary, and \textsc{gen}, featuring 1,492 synthetically generated usages of these slang terms, across diverse scenarios. We assess language models on three tasks: target word prediction (TWP), guided target word prediction (TWP$^*$) and target word selection (TWS). 
Our results reveal four key findings: (1) Higher average model performance TWS versus TWP and TWP$^*$, with average accuracy score increasing from 0.03 to 0.49 respectively (2) Stronger average model performance on \textsc{web} versus \textsc{gen} datasets, with average similarity score increasing by 0.03 and 0.05 across TWP and TWP$^*$ tasks respectively (3) en-IN tasks outperform en-AU when averaged across all models and datasets, with TWS demonstrating the largest disparity, increasing average accuracy from 0.44 to 0.54. These findings underscore fundamental asymmetries between generative and discriminative competencies for variety-specific language, particularly in the context of slang expressions despite being in a technologically rich language such as English.

\end{abstract}
\section{Introduction}
Varieties of a language are considered to differ in syntax, vocabulary and pragmatics~\cite{joshi-etal-2025-dialects}. A particularly unique aspect in terms of vocabulary is colloquial terms or slang. Slang is defined as language used by a particular group of people, during the period of its popularity~\cite{dumas-slang, slang-def}. In this paper, we investigate the ability of large language models (LLMs) to interpret slang in the context of Australian and Indian English, examples of which are as follows: 
\begin{description}
    \item[Prepone (Verb):] A colloquial Indian English term used to bring a scheduled event to an earlier time or date; the opposite of postpone. An example usage is `\textit{Since the manager is leaving early, we have decided to prepone the meeting to 10 AM.}'
    \item[Far Out (Adj):] A colloquial Australian English expression used to denote surprise, disbelief, or to describe something as being excellent or extreme. For example, `\textit{Far out, I can't believe how much that concert ticket cost!}'
\end{description}

Slang is inherently dynamic, community-specific, and culturally embedded, making it a critical starting point for evaluating whether contemporary language models can handle the full spectrum of linguistic diversity.
Recent research has highlighted systematic performance gaps when large language models (LLMs) process non-standard English varieties (\citealt{wuraola-etal-2024-understanding}, \citealt{khanuja-etal-2020-gluecos}, \citealt{deas-etal-2023-evaluation}), yet most work on slang interpretation has focused exclusively on Standard American English or has not distinguished between English varieties (\citealt{mei-etal-2024-slang}, \citealt{sun-etal-2024-toward}). This limitation overlooks the reality that slang terms are often variety-specific, carrying meanings and connotations that differ across geographical and cultural contexts. Being a global language, English has developed several colloqualisms in the geographies where it is spoken, highlighting the varied usage across cultures.

In this paper, we systematically evaluate how well state-of-the-art LLMs understand slang phrases from Indian English (en-IN) and Australian English (en-AU). The two are representative of two kinds of Englishes: en-IN may be spoken as an additional language while en-AU may be spoken as a first language. With these English varieties in focus, we address the research question: 

\emph{``How well do language models understand lexical variations in language varieties in terms of colloquial expressions or slang?''}
\begin{figure*}[t!]
    \centering
    \includegraphics[width=\linewidth]{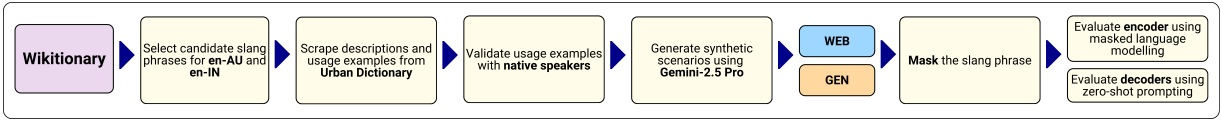}
    \caption{Methodological Overview of Our Approach.}
    \label{fig:method}
\end{figure*}

\begin{figure*}[t!]
    \centering
    \includegraphics[width=\linewidth]{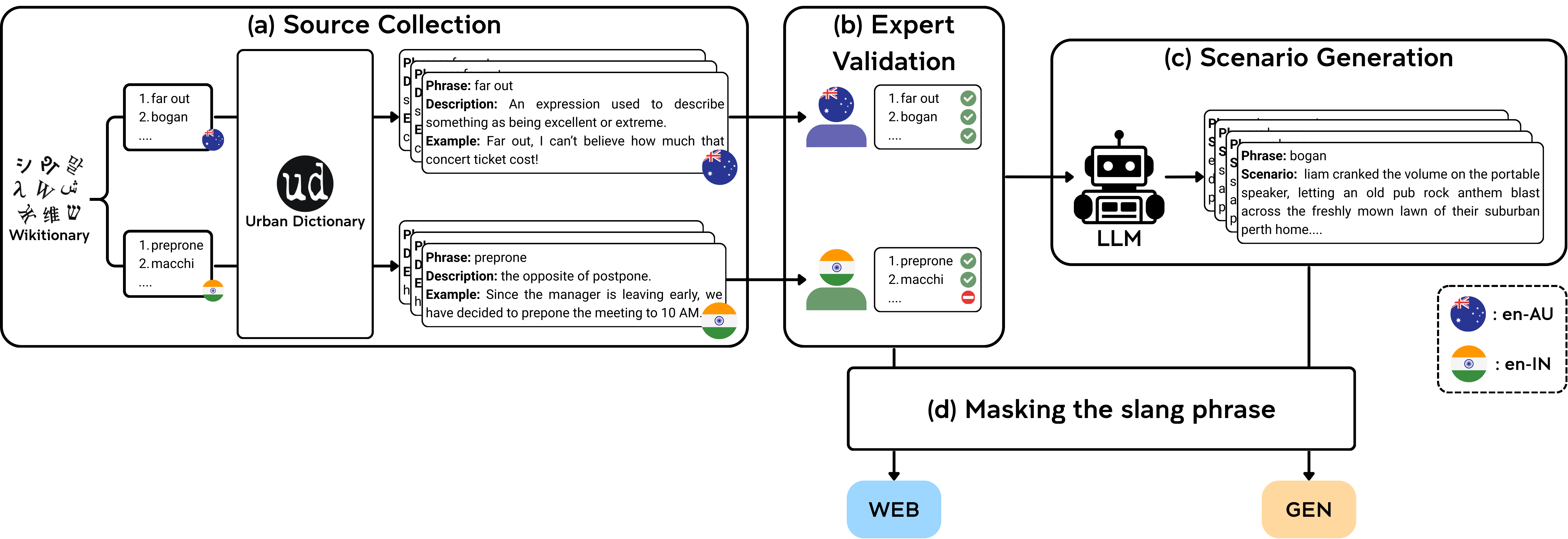}
    \caption{Dataset Creation.}
    \label{fig:data-create}
\end{figure*}

In order to do so, we set up two downstream tasks, inspired by~\citet{srirag-etal-2025-evaluating}: (a) \textbf{target word\footnote{Some slangs are phrases and are treated as such.} prediction}: Predicting a word in the place of a blanked position in an example sentence from among all words in the vocabulary, and (b) \textbf{target word selection}: Predicting a word in the place of a blanked position in an example sentence from among a set of options. We use example sentences from two sources: (a) slang definitions and example usages from Urban Dictionary\footnote{\url{www.urbandictionary.com}; Accessed on 31 December 2025.} that are manually curated by two native speakers; and (b) LLM-generated examples based on definitions from part (a). The sources result in two evaluative datasets: \textsc{web} and \textsc{gen}. We then evaluate several LLMs on target word prediction and target word selection. \textbf{Our evaluation highlights the limitations of LLMs to identify language variety-specific colloquial words in context, and bear implications to culturally aware NLP systems.}

This paper makes the following contributions:
\begin{enumerate}
    \item In the methodological sense, this is the first evaluative study that utilises target word selection and target word prediction using encoder and decoder models to evaluate if LLMs understand language variety-specific slangs, particularly for varieties of English.
    \item Novel Evaluation Datasets: Our datasets, \textsc{web}, containing web-sourced usage examples from Urban Dictionary, and \textsc{gen}, featuring synthetically generated diverse scenarios comprise 377 unique slang phrases across en-IN and en-AU varieties.
    \item Systematic Performance Analysis: We evaluate 7 language models ranging from 110M to 8B parameters, revealing key findings related to model performance in target word prediction and target word selection settings.
\end{enumerate}

Our findings demonstrate that while modern LLMs show promising capabilities compared to MLMs particularly for English, significant challenges remain in understanding variety-specific slang, such as Australian English. These results underscore the need for more inclusive training data and evaluation benchmarks that account for linguistic diversity beyond standard language varieties.

\section{Methodology}
We present a detailed overview of our evaluation methodology in Figure~\ref{fig:method}. Using web-based sources, we collect candidate slang phrases from the two varieties alongside their definitions and usages (\textsc{web}). We then manually validate the correctness using native speakers. We then use Google Gemini Pro 2.5 \cite{google_gemini_2025} to create scenarios and examples where such slang words will be used (\textsc{gen}). With the two datasets in place, we mask the slang phrase and get encoder and decoder models to predict the masked phrase.

\begin{table*}[t!]\centering
        \begin{tabular}{cccccccccc}
        \toprule
         \multirow{2}{*}{Variety} & \multicolumn{2}{c}{Count} & \multirow{2}{*}{$\bar{\text{p}}$} & \multirow{2}{*}{$\bar{\text{d}}$} & \multicolumn{2}{c}{$\bar{\text{u}}$} & \multicolumn{2}{c}{Perp.} & \multirow{2}{*}{Latest Entry}\\\cmidrule(lr){2-3}\cmidrule(lr){6-7}\cmidrule(lr){8-9}
         & \textsc{web} & \textsc{gen} & & & \textsc{web} & \textsc{gen} & \textsc{web} & \textsc{gen} & \\\midrule
         en-AU &  313 & 1244 & 7.7 & 138 & 90.4 & 443.0 & 701.5 & 48.9 &Sept 10 2025 \\
         en-IN & 64 & 248 & 8.0 & 197.7 & 114.1 & 468.2 & 406.5 & 47.1 & Oct 15 2025\\
        \bottomrule
    \end{tabular}
    \caption{Constructional statistics of \textsc{web} and \textsc{gen}. $\bar{\text{p}}$, $\bar{\text{d}}$ and $\bar{\text{u}}$ is the average {character} length of a slang phrase, corresponding definition and usage/scenario respectively. \textit{Perp.} represents average perplexity computed using GPT-2{~\cite{radford2019language}}.}
    \label{tab:data-stats}
\end{table*}

\begin{table}[t!]\centering
        \begin{tabular}{ccccc}
        \toprule
         Variety & R$_1$ & R$_2$ & R$_\text{L}$ & Sim.\\\midrule
         en-AU & 0.25 & 0.03 & 0.16 & 0.39\\
         en-IN & 0.24 & 0.03 & 0.15 & 0.37\\
        \bottomrule
    \end{tabular}
    \caption{Diversity analysis of generated scenarios in \textsc{gen}. R$_{\text{(1, 2, L)}}$ are average ROUGE scores and \textit{Sim.} is the average cosine similarity between scenario embeddings, extracted using allMiniLM-L6-v2{~\cite{reimers-gurevych-2019-sentence}}.}
    \label{tab:scenario-anal}
\end{table}

\subsection{Dataset Creation}
As described in Figure~\ref{fig:data-create}, we construct the dataset with English slang phrases used in two regions, hence covering two language varieties\footnote{We do not consider the subdialects present in both regions, and acknowledge the same as a reasonable limitation.}: Australian English (en-AU) and Indian English (en-IN). We collect the slang phrases from a web-based source and validate the relevance using native speakers. We also augment the dataset to include diverse scenarios with the usage of slang phrases.

\paragraph{Source Collection} We collate an initial list of the slang phrases, from both regions, using Wiktionary{\footnote{\url{en.wiktionary.org}; Accessed on 31 December 2025.}}. For each phrase, we extract corresponding definitions and usage examples from Urban Dictionary, a peer-contributed platform where users submit multiple definitions for individual phrases. Source Collection yields an initial list of 940 slang phrases for en-IN and 2540 slang phrases for en-AU. 


\paragraph{Expert Validation} We then employ one expert annotator from each region of interest to manually review and filter and remove irrelevant or incorrect entries. This process yields \textsc{web}, a high-quality subset of phrase, definition and usage example tuples from Urban Dictionary. Expert Validation removes 876 slang phrases from en-IN and 2227 slang phrases from en-AU. 

\paragraph{Scenario Generation} Using the phrases and their corresponding definitions from \textsc{web}, we prompt Google Gemini Pro 2.5 \cite{google_gemini_2025} to generate four unique scenarios that naturally motivate the usage of each phrase. The generation prompt (provided in Appendix \ref{app:scenario_gen}) instructs the model to create scenarios with named characters, specific settings, and single quotations containing the target phrase.

\subsection{Analysis}
Table \ref{tab:data-stats} provides a constructional statistics of both \textsc{web} and \textsc{gen}. We notice a higher number of slang phrases extracted for en-AU compared to en-IN. We also compute perplexity of the usage examples and generated scenarios, using GPT-2~\cite{radford2019language} from \textsc{web} and \textsc{gen} respectively. We observe a higher perplexity with the usage examples in \textsc{web} (en-AU: 701.5) as compared to the generated scenarios in \textsc{gen} (en-IN: 47.1). This is due to the usage examples in \textsc{web}, being derived from peer-contributed web content. Furthermore, both subsets differ in terms of character lengths, with \textsc{web} having shorter examples (en-IN: 114.1) compared to \textsc{gen} (en-AU: 443.0). Taken together, this disparity suggests that while \textsc{web} captures authentic naturalistic usage, it possesses lower overall linguistic fluency compared to the synthetically curated \textsc{gen}. \textbf{This observation is pertinent to future research in creation or collection of datasets for language varieties, particularly for high-resource languages such as English}. We present example slang phrases and their corresponding usage examples/scenarios in Appendix~\ref{app:data_examples}.

\paragraph{Scenario Diversity Analysis} We also evaluate the diversity of the generated scenarios in \textsc{gen} in terms of lexical overlap and semantic similarity. This evaluation is necessary because evaluating language models on similar scenarios, and usage examples does not yield any meaningful insights. For each slang phrase in \textsc{gen}, we compute a pairwise n-gram overlap between all the generated scenarios using ROUGE~\cite{lin-2004-rouge} to evaluate lexical diversity. Similarly, we compute pairwise cosine similarity between the scenario embeddings, extracted using allMiniLM-L6-v2~\cite{reimers-gurevych-2019-sentence}. Table \ref{tab:scenario-anal} reports low average semantic similarity, evidencing that the scenarios remain semantically dissimilar to each other, while also diverging lexically.

\subsection{Evaluative Tasks}

Given a usage example with the slang phrase, we mask the slang phrase. For inputs with multiple instances of the slang phrase, we create multiple instances with each position being masked. We use pre-trained language models to (a) predict the masked slang phrase; and (b) select the slang phrase from a set of options with other distractors. The two tasks are described as follows. 

\paragraph{Target Word Prediction (TWP)} In this task, the model is given a sentence in which a slang phrase has been masked and is asked to predict the missing phrase. The model is free to generate any word or phrase from its vocabulary, and the output is considered correct if it exactly matches the original slang phrase. With encoder-only language models, we predict the phrase at the masked position, utilising masked language modeling. For decoder-only large language models, we convert the masked sentence into a cloze-style prompt. An indicative prompt used for this task is provided in Appendix \ref{app:twp_prompt}. 

For decoder-only models, we additionally conduct a guided variant of the task (labeled TWP$^*$), where the prompt is augmented with an explicit instruction directing the model to generate a slang phrase suitable for the specific language variety. An indicative prompt used for this task is provided in Appendix \ref{app:twpg_prompt}. 

\paragraph{Target Word Selection (TWS)} We additionally evaluate models using a multiple-choice version of the task. The model is given a sentence with the slang phrase masked, along with a fixed set of candidate answers. The candidate set consists of four options: the correct slang phrase and three randomly selected distractor phrases drawn from the same language variety. The model is instructed to select the option that best fits the masked context. 

For decoder-only large language models, we add an explicit instruction: \textit{``Fill in the blank with the best-fitting answer.''}. An indicative prompt used for this task is provided in Appendix \ref{app:cloze_prompt}

\section{Experiment Setup}

We report the performance on \textit{seven} pre-trained language models including \textit{three} encoder models: BERT-Base (\textsc{bert}; \citealt{devlin-etal-2019-bert}), RoBERTa-Large (\textsc{roberta}; \citealt{liu2019robertarobustlyoptimizedbert}), XLM-RoBERTa-Large (\textsc{xlm}; \citealt{conneau-etal-2020-unsupervised}) and \textit{four} decoder models: Granite-4.0-1B (\textsc{granite}; \citealt{ibm2025granite4}), Llama-3.2-3B-Instruct (\textsc{llama}; \citealt{grattafiori2024llama3herdmodels}), Olmo-2-7B-Instruct (\textsc{olmo}; \citealt{olmo20252olmo2furious}), Qwen3-4b-Instruct (\textsc{qwen}; \citealt{yang2025qwen3technicalreport}). We evaluate encoder models under masked language modelling, where we extract the top-1 predictions from the masked position. For decoder models, we formulate the evaluation as a multiple-choice cloze format task with temperature$=$0.8. All experiments were conducted on an Apple M1 Pro with 16GB RAM using 8-bit quantized versions of the models to optimize inference efficiency on local hardware.

We report our results on two metrics: \textit{accuracy} and \textit{similarity}. \textit{Accuracy} is the proportion of instances where the model predicted the ground truth slang phrase. As similarity, we report the cosine similarity between the Sentence-BERT embeddings~\cite{reimers-gurevych-2019-sentence} of the reference slang phrase and the predicted slang phrase. To verify robustness of Sentence-BERT embeddings, we additionally compute cosine similarity between the Granite-embedding-125m-english embeddings \cite{awasthy2025graniteembeddingmodels} and the two embedding models' agreement by computing Pearsons Correlation. The average Pearson Correlation across all models, domains and varieties is 0.77. The complete results are tabulated in Appendix \ref{app:gen_full} and \ref{app:web_full}.

\section{Results} 

\begin{figure*}[t!]
    \centering
    \includegraphics[width=0.8\linewidth]{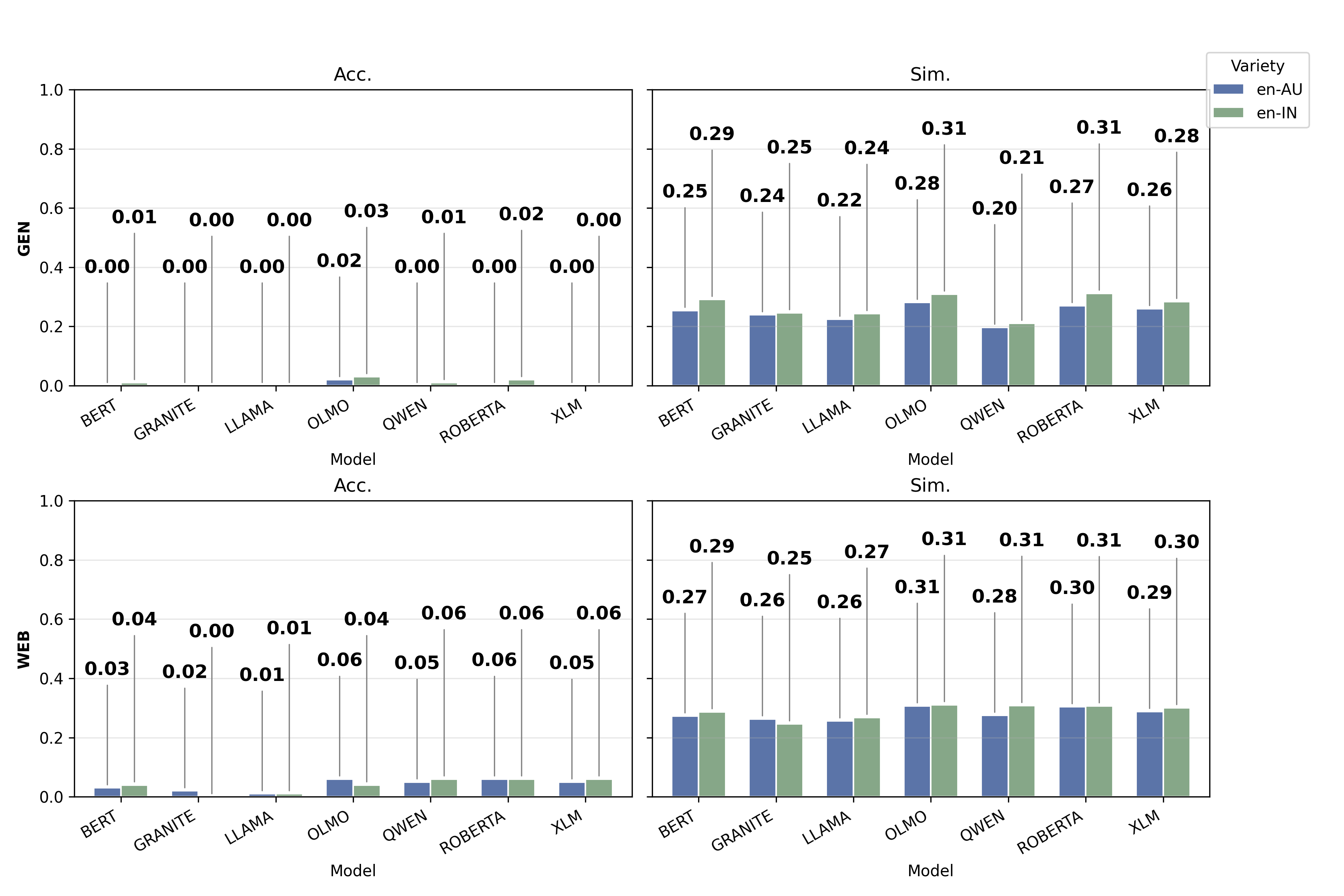}
    \caption{Performance comparison of various models on the target word prediction (TWP) task across en-AU and en-IN.}
    \label{fig:twp}
\end{figure*}

We present the results of experiments, centered around the following questions: (a) How well do language models perform on the downstream tasks? (Section~\ref{sec:overall}); (b) How do factors such as domain and language variety influence model performance? (Section~\ref{sec:factors}); (c) What types of errors do models make, and what do these errors reveal about model behaviour? (Section~\ref{sec:error-anal}).

\begin{figure*}[t!]
    \centering
    \includegraphics[width=0.8\linewidth]{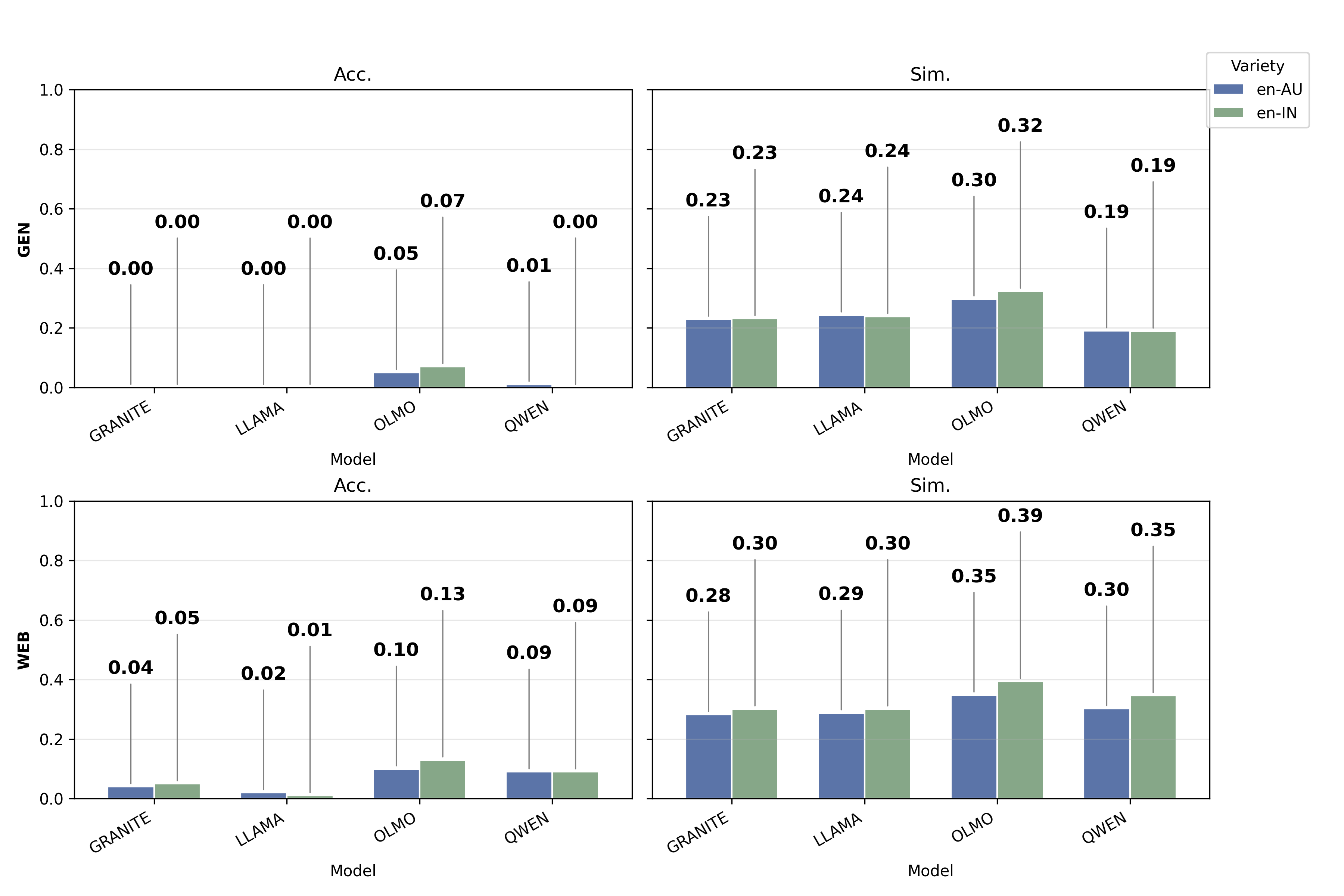}
    \caption{Performance comparison of various models on the guided target word prediction (TWP$^*$) task across en-AU and en-IN.}
    \label{fig:twpg}
\end{figure*}

\begin{figure*}[t!]
    \centering
    \includegraphics[width=0.8\linewidth]{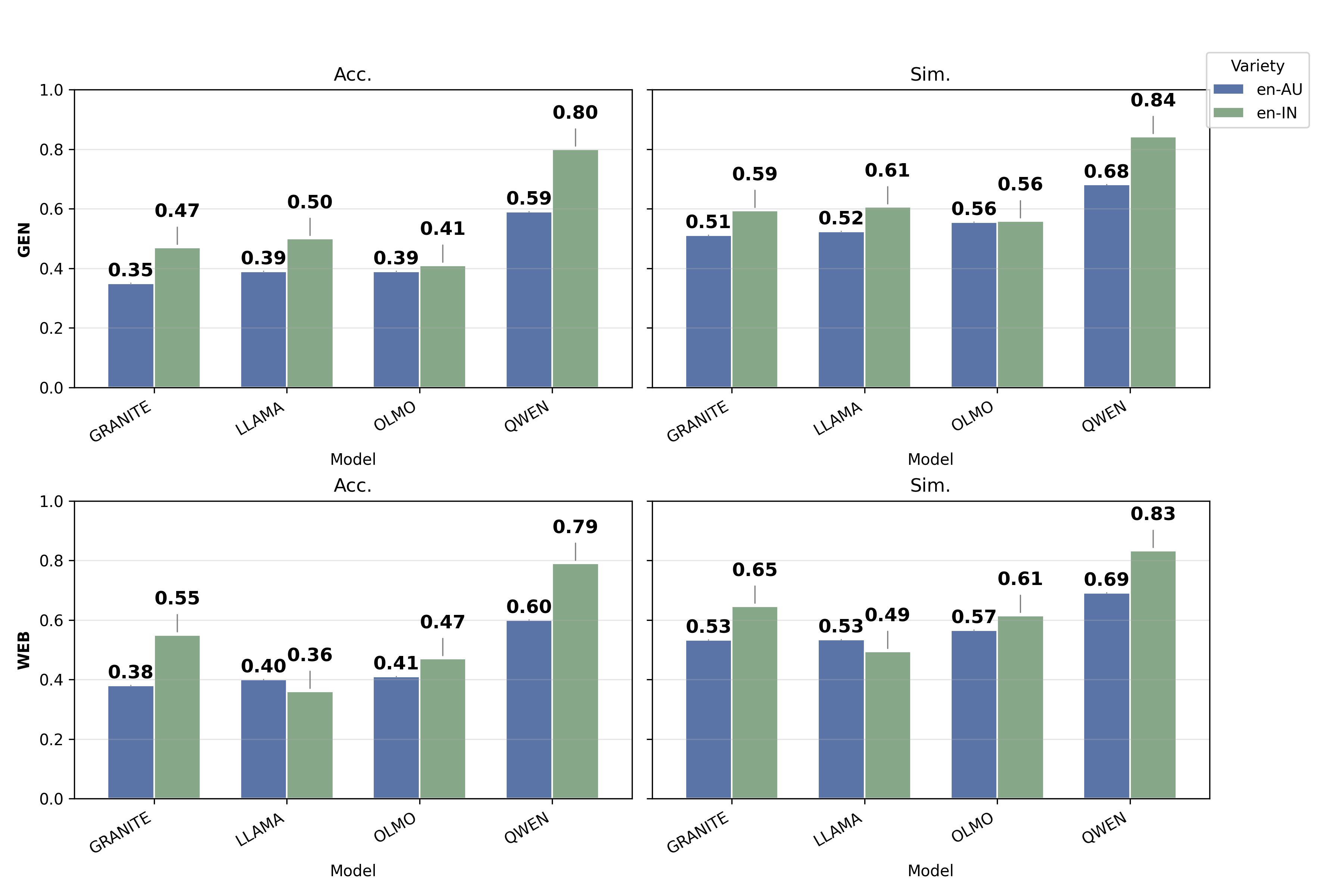}
    \caption{Performance comparison of various models on the target word selection (TWS) task across en-AU and en-IN.}
    \label{fig:tws}
\end{figure*}

\subsection{Task Performance}\label{sec:overall}

We first present the overall model performance on the two downstream tasks: target word prediction (TWP) and its guided variant (TWP$^*$) as well as target word selection (TWS). As shown in Table~\ref{tab:task_only_avg}, models perform poorly on TWP, achieving an average accuracy of 0.02 and an average similarity of 0.27. TWP$^*$ performs similarly poorly with an average accuracy of 0.04 and an average similarity of 0.28. This indicates that generating the correct slang phrase in an open-ended setting remains challenging. Figure~\ref{fig:twp} and Figure \ref{fig:twpg} further show that this difficulty is consistent across model architectures and language varieties, with only marginal differences between models and a maximum accuracy of 0.13. In contrast, models report higher performance on TWS. When the task is formulated as a multiple-choice cloze test, average accuracy increases to 0.49, with a corresponding similarity score of 0.61. This suggests that while current models struggle to generate slang expressions, they are more effective at recognising the correct phrase when the search space is constrained. Figure~\ref{fig:tws} shows that the improvement is consistent across model architectures and language varieties.

\subsection{Impact of Domain and Varieties}\label{sec:factors}

We next examine how domain and language variety influence model performance. Tables~\ref{tab:domain_effect} and~\ref{tab:variety_effect} summarise these effects by averaging results across all models.

\paragraph{Domain Effects.}
Table~\ref{tab:domain_effect} shows that models consistently perform better on data drawn from the \textsc{web} domain than on the \textsc{gen} domain, potentially due to data contamination. For TWP, web-based examples yield an accuracy of 0.04, compared to 0.01 on generated data. A similar drop is observed in semantic similarity. TWP$^*$ yields a larger drop, with accuracy decreasing from 0.07 to 0.02 across the sets. This gap is also present for TWS, where accuracy decreases from 0.50 on web data to 0.49 on generated data. These results indicate that domain shift has a measurable impact on the open-vocabulary tasks. Generated examples from \textsc{gen} appear to be more challenging, likely because they differ in style or contextual cues from naturally occurring web data. This observation resonates with past observations regarding data contamination and evaluation~\cite{dong-etal-2024-generalization}. Therefore, we recommend that \textbf{generating new data (either using a human or an LLM) is a more robust evaluation technique for language varieties.}

\paragraph{Language Variety Effects.}
Table~\ref{tab:variety_effect} reports performance differences between en-AU and en-IN. Across all tasks, models perform better on en-IN than on en-AU. For TWP, en-IN shows higher accuracy (0.03 vs. 0.02) and higher similarity (0.28 vs. 0.26). Models perform similarly across both metrics for TWP$^*$. The difference is even more pronounced for TWS, where accuracy improves by 0.10 and similarity by 0.08 when moving from en-AU to en-IN. This consistent improvement suggests that models are better aligned with slang usage patterns found in en-IN data, possibly due to greater representation or stylistic overlap in pre-training data. Importantly, the variety effect is larger than the domain effect for TWS, indicating that language variety plays a particularly significant role when models must discriminate between competing slang candidates.

\begin{table}[t!]
\centering
    \begin{tabular}{ccc}
        \toprule
        {Task} & {Acc.} & {Sim.} \\
        \midrule
        TWP & 0.02 & 0.27 \\
        TWP$^*$& 0.04 & 0.28 \\
        TWS  & 0.49 & 0.61 \\
        \bottomrule
    \end{tabular}
\caption{Average performance of language models across tasks, averaged over all models, varieties, and datasets.}
\label{tab:task_only_avg}
\end{table}

\begin{table*}[t]
\centering
    \begin{tabular}{ccccccc}
        \toprule
        \multirow{2}{*}{Domain} & \multicolumn{2}{c}{TWP} & \multicolumn{2}{c}{TWP$^*$} & \multicolumn{2}{c}{TWS} \\\cmidrule(lr){2-3} \cmidrule(lr){4-5} \cmidrule(lr){6-7}
        & Acc. & Sim. & Acc. & Sim. & Acc. & Sim.\\
        \midrule
        \textsc{web} & 0.04 & 0.29 &0.07&0.32& 0.50 & 0.61 \\
        \textsc{gen} & 0.01 \diff{0.03} & 0.26 \diff{0.03} &0.02 \diff{0.05}&0.24 \diff{0.08}& 0.49 \diff{0.01} & 0.61 \diff{0.00} \\
        \bottomrule
    \end{tabular}
\caption{Effect of domain on task performance, averaged across models and varieties. The decrease in performance is shown in \diff{red}.}
\label{tab:domain_effect}
\end{table*}

\begin{table*}[t]
\centering
    \begin{tabular}{ccccccc}
        \toprule
        \multirow{2}{*}{Variety} & \multicolumn{2}{c}{TWP} & \multicolumn{2}{c}{TWP$^*$} & \multicolumn{2}{c}{TWS} \\\cmidrule(lr){2-3} \cmidrule(lr){4-5}\cmidrule(lr){6-7}
         & Acc. & Sim. & Acc. & Sim.& Acc. & Sim. \\
        \midrule
        en-AU & 0.02 & 0.26 &0.04&0.27& 0.44 & 0.57 \\
        en-IN & 0.03 \improve{0.01} & 0.28 \improve{0.02} &0.05 \improve{0.01}&0.29 \improve{0.02}& 0.54 \improve{0.10} & 0.65 \improve{0.08} \\
        \bottomrule
    \end{tabular}
\caption{Effect of language variety on task performance, averaged across models and domains. The increase in performance is shown in \improve{blue}.}
\label{tab:variety_effect}
\end{table*}

\subsection{Error Analysis}\label{sec:error-anal}

To better understand the sources of model errors, we conduct a qualitative error analysis on the TWP task. We focus on this task because it exhibits the lowest average performance across all tasks, and therefore provides greater insight into model limitations. We select the best-performing model based on average similarity on the \textsc{gen} dataset, which is \textsc{olmo}. For each dataset (\textsc{web}, \textsc{gen}) and language variety (en-AU, en-IN), we extract the 30 lowest-scoring test instances according to semantic similarity, yielding a total of 120 examples. These examples are manually analysed by native speakers of the language varieties, and categorised into five recurrent failure types.

\paragraph{Error Categories} We briefly describe each error category below and provide illustrative examples in Table~\ref{tab:failure_examples}.
\begin{itemize}
    \item \textit{Literalisation} occurs when the model predicts a literal referent or standard lexical item corresponding to the slang phrase. While semantically appropriate, such predictions lose the idiomatic and regional character of the original expression.
    \item \textit{Generic Substitution} refers to cases where the model preserves the general tone or evaluative meaning but replaces the target phrase with a non-regional or broadly applicable alternative. This reflects limited sensitivity to locale-specific lexical choice.
    \item \textit{Semantic Drift} captures instances where the model remains within the correct topical or semantic field but selects a conceptually adjacent phrase that alters the intended meaning. These errors often arise in contexts that underspecify the precise pragmatic function of the slang term.
    \item \textit{Contextual Misinterpretation} occurs when the model fails to correctly interpret situational or discourse cues, resulting in a prediction that changes the narrative or pragmatic force of the sentence.
    \item \textit{True Failure} denotes outputs that are incoherent, grammatically ill-formed, or entirely unrelated to the surrounding context.
\end{itemize}

\begin{table*}[t!]
    \centering
    \begin{tabular}{p{3cm} p{2.2cm} p{2.6cm} p{2.5cm}}
        \toprule
        Category & Target phrase & Description & Prediction \\
        \midrule
        Literalisation &
        maggot bag &
        Australian slang for a meat pie &
        pie \\
        
        \midrule
        Generic Substitution & absolute unit &
        Someone very heavy or large &
        behemoth \\
        
        \midrule
        Semantic Drift &
        bogan &
        An unsophisticated person from a working-class background &
        heavy metal band \\
        
        \midrule
        Contextual Misinterpretation &
        a kangaroo loose in the top paddock &
        Intellectually inadequate &
        more to this picture \\
        
        \midrule
        True Failure &
        amber fluid &
        Australian slang for beer &
        34 \\
        \bottomrule
    \end{tabular}
    \caption{Representative examples of error categories observed in \textsc{olmo} predictions on the \textsc{gen} dataset.}
    \label{tab:failure_examples}
\end{table*}

\begin{table}[t!]
\centering
    \begin{adjustbox}{width=\linewidth}
        \begin{tabular}{lcccc}
            \toprule
            \multirow{2}{*}{Category} & \multicolumn{2}{c}{\textsc{gen}} & \multicolumn{2}{c}{\textsc{web}} \\
            \cmidrule(lr){2-3} \cmidrule(lr){4-5}
             & en-AU & en-IN & en-AU & en-IN \\
            \midrule[\heavyrulewidth]
            Literalisation & 2 & 3 & 3 & 5 \\
            Generic Substitution & 6 & 8 & 1 & 7 \\
            Semantic Drift & 7 & 6 & 3 & 4 \\
            Contextual Misinterpretation & 7 & 7 & 4 & 9 \\
            True Failure & 8 & 6 & 19 & 5 \\
            \hdashline
            Total & 30 & 30 & 30 & 30 \\
            \bottomrule
        \end{tabular}
    \end{adjustbox}
    \caption{Distribution of error categories across the 30 lowest-scoring instances per dataset and language variety for the best-performing model (\textsc{olmo}).}
    \label{tab:err_anal}
\end{table}

Table~\ref{tab:err_anal} summarises the distribution of error categories across datasets and language varieties. Across both datasets and varieties, the most frequent failure types are \textit{generic substitution}, \textit{semantic drift}, and \textit{contextual misinterpretation}. These categories reflect cases where the model captures aspects of the meaning or tone of the masked phrase but fails to recover the regionally appropriate slang expression. Notably, true failures, where the output is incoherent or unrelated, are relatively rare for en-IN but more prevalent for en-AU in the \textsc{web} dataset, suggesting uneven robustness across varieties and domains.

Overall, this analysis shows that most errors arise not from complete semantic failure, but from an inability to recover regionally appropriate slang expressions. Models frequently demonstrate partial understanding of meaning and tone, yet struggle with the cultural and pragmatic specificity required for accurate slang generation. This finding aligns with the large gap observed between similarity and accuracy in the TWP task.

\section{Related Work}
Recent work has increasingly recognised that LLMs exhibit systematic performance gaps when processing non-standard English varieties. \citet{wuraola-etal-2024-understanding} demonstrates that leading LLMs under-perform on comprehension tasks involving Nigerian English, particularly in emotion labeling and paraphrasing tasks. Similar studies on Indian English \cite{khanuja-etal-2020-gluecos}, African-American English \cite{deas-etal-2023-evaluation} and Nigerian English ~\cite{srirag2025predicting} reveal language models perform poorly in comparison to Standard American and British English varieties. These findings highlight that despite training on broad web corpora, contemporary language models fail to adequately represent linguistic diversity.

Slangs are a lexical evidence of the cultural knowledge of a community. Recent work explores whether language models capture cultural knowledge and conventions. \citet{seth-etal-2024-dosa} employ knowledge elicitation tasks to assess cultural familiarity across diverse contexts, finding that models underperform for non-Anglo-centric cultures. \citet{rao-etal-2025-normad} develop NormAd, a scenario-based evaluation framework for cultural norm understanding through closed-form QA, demonstrating that models struggle with non-Western cultural contexts. These works collectively demonstrate that language models' cultural knowledge remains predominantly Western-centric.

The focus of this paper is slang. Slang represents a particularly challenging domain for LLMs due to its dynamic, community-specific nature. Informal language exemplifies the transient nature of evolving languages; thus, language models trained on temporally fixed corpora raise questions about continuous adaptability to emerging linguistic concepts. \citet{mei-etal-2024-slang} propose a causal inference framework for slang comprehension using an Urban Dictionary-based dataset, demonstrating methods for adapting models to novel slang terms. \citet{sun-etal-2024-toward} compile a slang detection dataset from movie transcripts and achieve strong performance with both open and closed-source models on identification tasks. However, these works either focus exclusively on Standard American English or do not distinguish between English varieties, limiting their applicability to understanding variety-specific slang. Ours is the first work focusing on two varieties of English: Australian and Indian English which by themselves represent Englishes spoken as the first and additional language respectively.
\section{Conclusion}
This work presents the first systematic evaluation of language models' ability to understand variety-specific slang across Indian English (en-IN) and Australian English (en-AU). Through two complementary datasets; web-sourced examples (\textsc{web}) and synthetically generated scenarios (\textsc{gen}), we assessed seven language models ranging from 110M to 8B parameters using a mask-filling evaluation framework encompassing both target word prediction and target word selection tasks. Our findings revealed three key insights. First, all evaluated models—regardless of architecture or scale—struggle to spontaneously generate appropriate slang terms in context, achieving at most 0.13 average accuracy in open-vocabulary prediction. This suggests that slang comprehension remains a significant blind spot even for state-of-the-art systems. Second, models show dramatic improvement when the task is reformulated as multiple-choice selection (with average accuracies reaching up to 0.8), this success highlights a fundamental asymmetry: models can discriminate between plausible slang alternatives more effectively than they can generate contextually appropriate slang. Third, larger models exhibit greater sensitivity to distributional shifts between naturalistic and synthetic contexts, with performance gaps of up to 0.1, suggesting that increased scale may lead to overfitting on specific textual patterns rather than robust understanding of lexical variation.

Our systematic failure analysis of 120 worst-performing predictions reveals the underlying nature of these limitations. We identify five distinct error categories that illuminate how models fail: literalization, generic substitution, semantic drift, contextual misinterpretation, and true failure. Critically, the prevalance of literalization and generic substitution errors reveal that models understand the semantic content but fail to recognize or generate the culturally embedded linguistic forms that distinguish regional varieties. The prevalence of these errors demonstrates that models have learned to map slang to underlying concepts but do not appropriately deploy variety-specific expressions. 
These results have important implications for the development of language technologies that serve diverse global populations. As language models are increasingly deployed in applications ranging from content moderation to educational tools, their limitations in understanding non-standard varieties and informal language pose risks of bias and reduced utility for speakers of these varieties. Our datasets and evaluation framework provide a foundation for future research on variety-specific informal language understanding, enabling more comprehensive assessments of linguistic diversity in natural language processing systems. Future work should explore methods for improving slang comprehension, including targeted data collection, variety-aware training objectives, and continual learning approaches that can adapt to evolving linguistic phenomena.

\section*{Limitations}
While we introduce two novel datasets, the \textsc{web} component is relatively small, comprising 377 unique entries. Furthermore, these examples are sourced exclusively from Wiktionary and Urban Dictionary. While these platforms provide valuable peer-contributed data, they may contain noise or demographic biases that do not fully reflect the breadth of spoken slang in these regions. Secondly, our analysis is restricted to two specific varieties: Indian English (en-IN) and Australian English (en-AU). While these varieties offer a comparison between two English typologies, our observations regarding model scaling behaviors and performance gaps may not generalize to other non-standard varieties, such as African American Vernacular English or Nigerian English. Finally, our dataset curation process relied on a single expert annotator per variety to validate the examples. Although this ensures native-level verification, it precludes the calculation of inter-annotator agreement metrics and leaves the dataset potentially susceptible to individual biases. Similarly, observations on the \textsc{web} dataset likely indicate data contamination.
\section*{Ethical Considerations}
As the regulation corpus was sourced using publicly available repositories, there are no significant ethical considerations to report. 

\section*{Acknowledgment}
This work was funded by Google’s ExploreCSR grant, awarded to Aditya Joshi in 2024.
\bibliography{custom}

\appendix
\section{Novel Scenario Generation Prompt}
\label{app:scenario_gen}
\texttt{Given a slang phrase and its definition, create four unique scenarios. Each scenario must introduce named characters, describe the setting, and contain a single quotation that uses the word exactly once. Do not include any titles, scenario numbers, labels, emojis, or explanatory text; only output the four resulting paragraphs, separated by a line break.}
\newline
\texttt{phrase: \{\textbf{phrase}\}}
\newline
\texttt{definition: \{\textbf{definition}\}}
\section{Dataset Examples}
Refer to Table \ref{app:data_examples_table} for data set examples.
\label{app:data_examples}
\begin{table*}[h!]
\small
\centering
\begin{tabular}{p{2.5cm} p{6cm} p{6cm}}
\toprule
Variety & en-AU & en-IN \\
\midrule
phrase $p$        & smoko & prepone \\
\hline
definition $d$  & a slang term used on building sites in Australia, meaning a morning-tea break, or a smoke break. & Function: transitive verb

Inflected forms: pre•poned; pre•pon•ing

Etymology: Latin preponere to place before, prepone, from pre- + ponere to place -- more at POSITION

Date: Has been in use in urban English spoken in India since at least the 1950s

To advance an event or activity to an earlier time. The closest American usage is ``to advance'' the timing of something. The word came into vogue in urban India as the opposite of ``postponing'' something. \\
\midrule
usage example $u$     & ``we'll knock off at 11 for smoko'' &
``To make sure we get to enjoy the fireworks display that starts promptly at 9 PM, let us prepone the dinner engagement to 7 rather than 8 tomorrow evening'' \\
\midrule
usage scenario $u$   & The fluorescent lights of the London advertising agency hummed, a stark contrast to the quiet focus on Anika's face as she stared at the mock-up on her screen. Her colleague, Ben, who had recently transferred from their Melbourne office, spun around in his chair. "You've been at that for hours, I'm grabbing a coffee and a biscuit, you keen for a quick smoko?" & Anjali tapped her stylus against the glass wall of the conference room in their bustling Bangalore office, catching Rohan's attention as he walked by. The quarterly progress charts were displayed on the large monitor, but her focus was clearly elsewhere. "Rohan, I just got off the phone with the clients from Singapore; they're flying in two days earlier than planned, so we need to prepone the final project presentation to Tuesday morning." \\
\bottomrule
\end{tabular}
\caption{Examples from \textsc{web} and \textsc{gen} datasets. }
\label{app:data_examples_table}
\end{table*}
\section{Target Word Prediction Prompt Example}
\label{app:twp_prompt}
\texttt{Fill in the blank with the best-fitting answer.
}
\newline
\texttt{Sentence: i'd rather ride on the \_\_\_\_ as opposed to taking the stairs.}
\newline
\texttt{Answer (return only the answer with no extra text): }
\section{Guided Target Word Prediction Prompt Example}
\label{app:twpg_prompt}
\texttt{Fill in the blank with the best-fitting answer.
}
\newline
\texttt{The answer is a Australian English slang word or phrase.
}
\newline
\texttt{Sentence: head on down to the beach this sunday, theres gonna be a \_\_\_\_ for the homeless.
}
\newline
\texttt{Answer (return only the answer with no extra text): 
}
\section{Target Word Selection Prompt Example}
\label{app:cloze_prompt}
\texttt{Fill in the blank with the best-fitting answer.}
\newline
\texttt{Sentence: "what's up jake?"
"i just got bitten by a \_\_\_\_."}
\newline
\texttt{Options: [mossie, gonski, ripping, munted]}
\newline
\texttt{Answer (return only the text of one option with no extra text)}
\section{Experimental Language Models Statistics}
\label{app:experimental_models}
Refer to Table \ref{app:experimental_models_table} for experimental language model's statistics.

 \begin{table}[!ht]
     \centering
     \begin{tabular}{p{3cm} p{1.5cm} l}
\toprule
         Language Model & Parameters & Release Date \\
\midrule
         bert-base-uncased & 110M & October 2018 \\
         roberta-large & 355M & July 2019 \\
         xlm-roberta-large & 550M & November 2019 \\
         granite-4.0-1b& 1B & October 2025 \\
         llama-3.2-3b-instruct & 3B & September 2024 \\
         olmo-2-1124-7b-instruct & 7B & November 2024 \\
         qwen3-4b-instruct-2507 & 4B & July 2025 \\
         \bottomrule
     \end{tabular}
     \caption{Language Models used in Experimental Setup, their approximate Parameter Counts and Public Release Date}
     \label{app:experimental_models_table}

 \end{table}
 \section{Model Performance Metrics With the Novel Scenario Dataset \textsc{gen}}
 \label{app:gen_full}
Refer to Table \ref{app:gen_full_results} for model's individual results using the \textsc{gen} dataset.
 \begin{table*}[!ht]
 \centering
\begin{tabular}{lcrrrrrr}
\toprule
Language Model & Variety & Total & EM & Acc. & Sim (G) & Sim (M) & Corr \\
\midrule
\multicolumn{8}{c}{\textbf{Target Word Prediction}} \\
bert-base-uncased & en-AU & 1238 & 2 & 0.002 & 0.726 & 0.254 & 0.341 \\
bert-base-uncased & en-IN & 243 & 3 & 0.012 & 0.729 & 0.291 & 0.670 \\
roberta-large & en-AU & 1238 & 5 & 0.004 & 0.730 & 0.270 & 0.464 \\
roberta-large & en-IN & 243 & 6 & 0.025 & 0.738 & 0.312 & 0.742 \\
xlm-roberta-large & en-AU & 1238 & 3 & 0.002 & 0.726 & 0.260 & 0.370 \\
xlm-roberta-large & en-IN & 243 & 0 & 0.000 & 0.727 & 0.284 & 0.494 \\
granite-4.0-1b-Q8\_0 & en-AU & 1232 & 1 & 0.001 & 0.729 & 0.239 & 0.307 \\
granite-4.0-1b-Q8\_0 & en-IN & 245 & 1 & 0.004 & 0.729 & 0.246 & 0.457 \\
Llama-3.2-3B-Instruct-Q8\_0 & en-AU & 1232 & 0 & 0.000 & 0.737 & 0.224 & 0.504 \\
Llama-3.2-3B-Instruct-Q8\_0 & en-IN & 245 & 1 & 0.004 & 0.733 & 0.243 & 0.716 \\
olmo-2-1124-7B-instruct-Q8\_0 & en-AU & 1232 & 23 & 0.019 & 0.741 & 0.281 & 0.713 \\
olmo-2-1124-7B-instruct-Q8\_0 & en-IN & 245 & 8 & 0.033 & 0.744 & 0.309 & 0.733 \\
Qwen3-4B-Instruct-2507-Q8\_0 & en-AU & 1232 & 3 & 0.002 & 0.729 & 0.197 & 0.482 \\
Qwen3-4B-Instruct-2507-Q8\_0 & en-IN & 245 & 3 & 0.012 & 0.732 & 0.210 & 0.617 \\
\midrule
\multicolumn{8}{c}{\textbf{Guided Target Word Prediction}} \\
granite-4.0-1b-Q8\_0          & en-AU & 1232 & 3  & 0.002  & 0.728 & 0.229 & 0.439 \\
granite-4.0-1b-Q8\_0          & en-IN & 245  & 1  & 0.004  & 0.727 & 0.231 & 0.436 \\
Llama-3.2-3B-Instruct-Q8\_0   & en-AU & 1232 & 6  & 0.005  & 0.740 & 0.243 & 0.701 \\
Llama-3.2-3B-Instruct-Q8\_0   & en-IN & 245  & 1  & 0.004  & 0.731 & 0.238 & 0.754 \\
olmo-2-1124-7B-instruct-Q8\_0 & en-AU & 1232 & 60 & 0.049  & 0.752 & 0.297 & 0.845 \\
olmo-2-1124-7B-instruct-Q8\_0 & en-IN & 245  & 17 & 0.069  & 0.757 & 0.323 & 0.816 \\
Qwen3-4B-Instruct-2507-Q8\_0  & en-AU & 1232 & 10 & 0.008  & 0.728 & 0.190 & 0.616 \\
Qwen3-4B-Instruct-2507-Q8\_0  & en-IN & 245  & 0  & 0.000  & 0.724 & 0.189 & 0.419 \\
\midrule
\multicolumn{8}{c}{\textbf{Target Word Selection}} \\
granite-4.0-1b-Q8\_0 & en-AU & 1232 & 437 & 0.355 & 0.828 & 0.511 & 0.978 \\
granite-4.0-1b-Q8\_0 & en-IN & 245 & 116 & 0.474 & 0.857 & 0.594 & 0.981 \\
Llama-3.2-3B-Instruct-Q8\_0 & en-AU & 1232 & 483 & 0.392 & 0.837 & 0.524 & 0.978 \\
Llama-3.2-3B-Instruct-Q8\_0 & en-IN & 245 & 123 & 0.502 & 0.863 & 0.606 & 0.985 \\
olmo-2-1124-7B-instruct-Q8\_0 & en-AU & 1232 & 485 & 0.394 & 0.839 & 0.555 & 0.982 \\
olmo-2-1124-7B-instruct-Q8\_0 & en-IN & 245 & 100 & 0.408 & 0.839 & 0.559 & 0.981 \\
Qwen3-4B-Instruct-2507-Q8\_0 & en-AU & 1232 & 725 & 0.589 & 0.890 & 0.681 & 0.987 \\
Qwen3-4B-Instruct-2507-Q8\_0 & en-IN & 245 & 196 & 0.800 & 0.946 & 0.842 & 0.987 \\
\bottomrule
\end{tabular}%
\caption{Model performance metrics with \textsc{gen} across Target Word Prediction, Guided Target Word Prediction and Target Word Selection. We provide results on Exact Matches (EM), Accuracy (Acc), Sim(G) and Sim(M) refer to average cosine similarity scores using embedding models Granite-embedding-125m-english embeddings and \cite{awasthy2025graniteembeddingmodels} Sentence-BERT embeddings~\cite{reimers-gurevych-2019-sentence} respectively. We compute Pearson Correlation (Corr.) between the resulting similarity scores to measure the two models agreement.}
  \label{app:gen_full_results}
\end{table*}

\section{Model Performance Metrics With the Web-Based Example Dataset \textsc{web}}
\label{app:web_full}
Refer to Table \ref{app:web_full_results} for model's individual results using the \textsc{web} dataset.
\begin{table*}[!ht]
\centering
\begin{tabular}{lcrrrrrr}
\toprule
Language Model & Variety & Total & EM & Acc. & Sim (G) & Sim (M) & Corr \\
\midrule
\multicolumn{8}{c}{\textbf{Target Word Prediction}} \\
bert-base-uncased & en-AU & 375 & 11 & 0.029 & 0.736 & 0.273 & 0.777 \\
bert-base-uncased & en-IN & 77 & 3 & 0.039 & 0.737 & 0.287 & 0.796 \\
roberta-large & en-AU & 375 & 22 & 0.059 & 0.743 & 0.304 & 0.824 \\
roberta-large & en-IN & 77 & 5 & 0.065 & 0.747 & 0.307 & 0.796 \\
xlm-roberta-large & en-AU & 375 & 17 & 0.045 & 0.737 & 0.288 & 0.831 \\
xlm-roberta-large & en-IN & 77 & 5 & 0.065 & 0.741 & 0.301 & 0.861 \\
granite-4.0-1b-Q8\_0 & en-AU & 374 & 6 & 0.016 & 0.737 & 0.263 & 0.700 \\
granite-4.0-1b-Q8\_0 & en-IN & 77 & 0 & 0.000 & 0.724 & 0.246 & 0.385 \\
Llama-3.2-3B-Instruct-Q8\_0 & en-AU & 374 & 3 & 0.008 & 0.738 & 0.256 & 0.673 \\
Llama-3.2-3B-Instruct-Q8\_0 & en-IN & 77 & 1 & 0.013 & 0.734 & 0.268 & 0.732 \\
olmo-2-1124-7B-instruct-Q8\_0 & en-AU & 374 & 24 & 0.064 & 0.748 & 0.307 & 0.836 \\
olmo-2-1124-7B-instruct-Q8\_0 & en-IN & 77 & 3 & 0.039 & 0.744 & 0.311 & 0.822 \\
Qwen3-4B-Instruct-2507-Q8\_0 & en-AU & 374 & 17 & 0.046 & 0.745 & 0.275 & 0.851 \\
Qwen3-4B-Instruct-2507-Q8\_0 & en-IN & 77 & 5 & 0.065 & 0.757 & 0.308 & 0.884 \\
\midrule
\multicolumn{8}{c}{\textbf{Guided Target Word Prediction}} \\
granite-4.0-1b-Q8\_0          & en-AU & 374  & 16 & 0.043  & 0.741 & 0.282 & 0.853 \\
granite-4.0-1b-Q8\_0          & en-IN & 77   & 4  & 0.052  & 0.744 & 0.301 & 0.849 \\
Llama-3.2-3B-Instruct-Q8\_0   & en-AU & 374  & 8  & 0.021  & 0.749 & 0.288 & 0.750 \\
Llama-3.2-3B-Instruct-Q8\_0   & en-IN & 77   & 1  & 0.013  & 0.736 & 0.301 & 0.748 \\
olmo-2-1124-7B-instruct-Q8\_0 & en-AU & 374  & 38 & 0.102 & 0.768 & 0.348 & 0.893 \\
olmo-2-1124-7B-instruct-Q8\_0 & en-IN & 77   & 10 & 0.130 & 0.774 & 0.394 & 0.889 \\
Qwen3-4B-Instruct-2507-Q8\_0  & en-AU & 374  & 34 & 0.091  & 0.760 & 0.302 & 0.888 \\
Qwen3-4B-Instruct-2507-Q8\_0  & en-IN & 77   & 7  & 0.091  & 0.759 & 0.346 & 0.899\\
\midrule
\multicolumn{8}{c}{\textbf{Target Word Selection}} \\
granite-4.0-1b-Q8\_0 & en-AU & 374 & 142 & 0.380 & 0.836 & 0.533 & 0.978 \\
granite-4.0-1b-Q8\_0 & en-IN & 77 & 42 & 0.546 & 0.873 & 0.646 & 0.990 \\
Llama-3.2-3B-Instruct-Q8\_0 & en-AU & 374 & 151 & 0.404 & 0.842 & 0.534 & 0.976 \\
Llama-3.2-3B-Instruct-Q8\_0 & en-IN & 77 & 28 & 0.364 & 0.824 & 0.494 & 0.988 \\
olmo-2-1124-7B-instruct-Q8\_0 & en-AU & 374 & 155 & 0.414 & 0.846 & 0.566 & 0.980 \\
olmo-2-1124-7B-instruct-Q8\_0 & en-IN & 77 & 36 & 0.468 & 0.855 & 0.615 & 0.981 \\
Qwen3-4B-Instruct-2507-Q8\_0 & en-AU & 374 & 225 & 0.602 & 0.893 & 0.691 & 0.987 \\
Qwen3-4B-Instruct-2507-Q8\_0 & en-IN & 77 & 61 & 0.792 & 0.942 & 0.833 & 0.991 \\
\bottomrule
\end{tabular}%
\caption{Model performance metrics with \textsc{web} across Target Word Prediction, Guided Target Word Prediction and Target Word Selection. We provide results on Exact Matches (EM), Accuracy (Acc), Sim(G) and Sim(M) refer to average cossine similarity scores using embedding models Granite-embedding-125m-english embeddings and \cite{awasthy2025graniteembeddingmodels} Sentence-BERT embeddings~\cite{reimers-gurevych-2019-sentence} respectively. We compute Pearson Correlation (Corr) between the resulting similarity scores to measure the two models agreement.}
\label{app:web_full_results}
\end{table*}
\end{document}